# Attention-Based Spatio-Temporal Approach for Fire and Smoke Detection

A. Mirrashid, M. Khoshbin, A. Atghaei, H. Shahbazi

***Abstract*—**In various industries, smoke and fire are two of the most important threats in the workplace. One of the common methods for detecting smoke and fire is the use of infrared thermal and smoke sensors, which cannot be used in outdoor applications. Therefore, the use of vision-based methods seems necessary. The problem of smoke and fire detection is spatiotemporal and requires spatiotemporal solutions. This paper presents a method that uses spatial features along with temporal-based features to detect smoke and fire in the scene. It consists of three main parts; the task of each part is to reduce the error of the previous part so that the final model has a robust performance. This method also uses transformer modules to increase the accuracy of the model. The results of our model show the proper performance of the proposed approach in solving the problem of smoke and fire detection and can be used to increase workplace safety.

***Keywords*—**Attention, fire detection, smoke detection, spatiotemporal.

## I. Introduction

With the expansion of industrial environments, the detection of smoke and fire in its early stages, especially in areas where it is not possible to install thermal sensors is necessary to mitigate fire and smoke threads and ensure the environment's safety.

Older methods of detecting smoke and fire were based on color, texture, and motion characteristics. Despite the simplicity and high speed of these algorithms and the lack of need for a lot of data to design them, their operational problems such as false alarms or high miss detection, make them not practical to use.

Recently, with the development of deep learning methods and the expansion of available data, deep learning-based methods [1]-[4] have been used widely. Most of these methods decide on the presence or absence of smoke and fire based on the image, which makes it impossible to properly distinguish smoke and fire from similar objects such as the sun and street lights. Therefore, in order to be able to better detect and reduce the error of the detection model, it is necessary to study the behavior in the time domain in addition to appearance features. The problem in this area is the lack of proper data and the variety and low speed of models with video input, which in practice makes them difficult to use.

In this paper, the smoke and fire detection process is designed in such a way that the final model has high speed and accuracy. For this purpose, a large dataset of images and videos related to smoke and fire has been collected and the models have been trained based on this dataset. Also, the module placement process is such that each model simultaneously reduces the computational complexity of the next model and its errors so that the final model is operational. The method presented in this paper is a combination of the methods mentioned. In a way, conventional image processing algorithms are used first to determine the candidate areas of smoke and fire, and then these areas are identified as input to the model to detect smoke and fires in these areas. Finally, in order to reduce the false alarm error of the detection model, a spatiotemporal model is applied to the sequence of candidate frames resulting from the previous two modules to correct its class. Also, the model is presented in such a way that it has the ability to detect smoke and fire with different dimensions and shapes.

The main contributions of this paper are as follows:

1) Considering both the chromatic and dynamic features of fire and smoke in the scene
2) Building a module to reduce errors and increase model decision reliability using transformers.
3) Reducing the computational complexities of the model by using three different modules.

This paper is organized as follows: Section II briefly reviewed some previous related works. The proposed method is presented in Section III. In Section IV, experimental results of the model are provided. Finally, the conclusion of this paper is presented in Section V.

## II. Related Works

There are several ways to detect smoke and fire. These methods can be classified into two main categories: 1) based on sensors and 2) based on vision. The focus of this article is on vision-based methods, which are especially necessary for outdoor environments where it is not possible to use sensors. Vision-based methods can be divided into two categories. The first category uses feature extraction and machine learning methods and the second category uses deep neural networks.

### *A. Methods Based on Feature Extraction and Machine Learning*

In this case, to detect smoke and fire, the feature vector is first calculated based on the user's desired features. These features include color, motion, optical flow, and the shape of objects in the image. The calculated features are then given to a decision algorithm to decide whether or not there is smoke or fire in the image.

In [5] a method for fire detection using color and motion characteristics is presented. In this paper, in addition to common color and motion features, wavelet transform is used

A. Mirrashid is with the VEUNEX, Germany (corresponding author, e-mail: a.mirrashid@veunex.com).

M. Khoshbin, A. Atghaei and H. Shahbazi are with the VEUNEX, Germany.

to analyze behavior and extract features in videos. This method requires thresholding to detect fire candidate areas. A method based on color and motion characteristic analysis for smoke and fire detection is presented in [6]. The method they proposed in [6] is to use some thresholds on RGB and HIS (hue, intensity, saturation) values and another threshold for motion detection based on the changes in the color of pixels in time. In [7] image processing is used to detect fires in forest areas. In this work, an algorithm based on YCbCr color space (Y is luma (brightness), Cb is blue minus luma (B-Y) and Cr is red minus luma (R-Y)) in addition to RGB values is presented to increase the accuracy of detecting fire areas. Reference [8] uses color and motion properties to detect smoke and fire simultaneously. In this work, fuzzy rules are used to improve classification performance. In [9], color and motion properties have also been used to detect smoke and fire. An algorithm for smoke detection using moving smoke is given in [10]. The method presented in [11] uses the edge detection algorithm to detect fire. In [12] the dynamic properties of fire are used to identify areas of fire, but in cases where objects similar to fire are present in the image, the performance of their method will be degraded. The authors of [13] have used two optical flow branches to distinguish fire areas from other areas. In [14] fire detection in forest areas has been investigated. This was done by converting the image to non-overlapping patches and then categorizing each patch using the AlexNet [15] model to fire and not fire. In [16] static and dynamic features are used simultaneously to detect fire.

The advantage of these methods is that they do not require a lot of data. Also, by considering the necessity of movement, the classification of items such as the sun as fire is prevented. The disadvantage of these methods is that due to the extraction of features based on items such as color, the error of these algorithms is very high, for example, items such as a moving orange box, will be detected as fire. Another problem with the methods presented in this field is the need to adjust the relevant thresholds, which is a time-consuming task with a high false alarm. On the other hand, feature design requires a lot of experience in order to design suitable features.

### *B. Methods Based on Deep Neural Networks*

Recently, the use of deep learning methods to detect the presence of smoke or fire in images has become widespread. Artificial intelligence-based methods have solved the mentioned disadvantages of feature-based methods. Most deep-learning smoke and fire detection methods fall into two main categories:

1) Classification-based methods: In these methods, the whole image has a label that indicates fire, smoke, or background. The main problem with this method is that in most cases, the ratio of smoke and fire size to the original image size is small, and using this method can cause false alarms or miss detection based on the threshold value. The advantage of these methods is that it is easier to create data for them. The lack of shape in smoke and fire makes it difficult to create labeled data for the detection and segmentation of fire and smoke. In [17] and [18] they have used the rearrangement of common classification models for fire detection. In [19] a hybrid model uses the Adaboost-LBP [20] model to create candidate areas and finally a convolutional neural network to detect fire. Fire detection using EfficientNet [21] architecture is given in [22]. In this architecture, the focal cost function is used to solve the problem of unbalanced categories [23]. An in-depth learning method has been developed to detect smoke and fire using images received from the camera. A normalized 14-layer convolutional architecture for fire feature and classification is presented in [24]. In [25], convolution with specific strides is used to detect fire and it is shown that the performance of the proposed model works better than previous models using fine-tuning of pre-trained networks.
2) Detection-based methods: In these methods, using detection models, in addition to classification, the location of smoke and fire is also determined and these approaches have the ability to detect different dimensions of smoke and fire. The problem with this method is the difficulty of creating a labeled dataset for it. A convolutional architecture that is derived from SqueezeNet for fire detection is presented in [26]. In this method, in order to reduce the calculations and execution speed, the dimensions of the filters in convolutional layers have been reduced and their proposed architecture has not any fully connected layers. Reference [27] presents an automatic method for detecting fire pixels without the use of time information. The possibility of implementing an artificial intelligence model on devices with limited computing power is presented in [28], in which a lightweight model for fire detection is presented. The method presented in [29] uses the common YOLO architecture to detect fire and its location in images. The faster RCNN architecture in [30] has been used to detect fire areas.

The main problem with AI-based methods is the need for a lot of training data and a time-consuming training process along with a lack of control over the smoke and fire detection process. This issue is seen more due to the lack of large and standard datasets with suitable variety. In this article, for proper learning, a wide variety of datasets are provided to solve these problems.

## III. Proposed Method

The model presented in this paper, as shown in Fig. 1, consists of 3 main parts: 1) the task of the first part is to find candidate areas for smoke and fire, 2) the second part is the smoke and fire detection model and 3) the third part is the final video-based classification model. In the first part, color and motion-based algorithms are used to identify candidate areas, the purpose of which is to reduce the processing of subsequent modules and reduce their error. The input of this module is a set of frames and its output is a binary map in which the pixels that are candidates for smoke or fire are 1 and the rest of the pixels are 0. In the second part, a detection network is used to detect the smoke and fire areas in the image. The first input of this network is the original image and the second input is the output of the previous module, which is used to cut the candidate part of the original image at the input of the network. Finally, in order to reduce the error caused by the first two modules and to investigate the spatiotemporal behavior of the smoke and fire candidate areas, a video-based network is used. The input of this module is a set of frames and the output of the second module. In fact, the purpose of this module is to decide on the category of extracted parts of the second module in a sequence of frames. In the following, each of the three modules is explained in more detail.

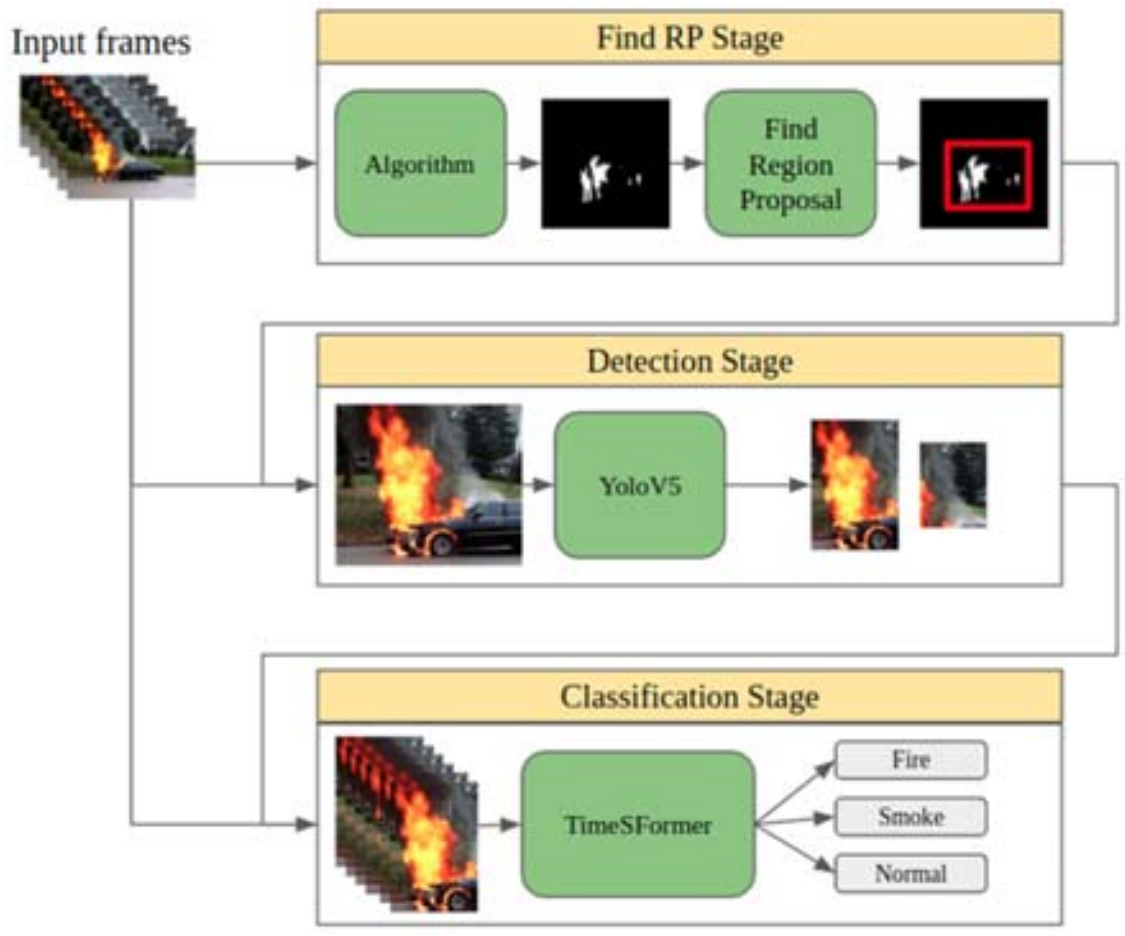

Fig. 1 Our proposed approach containing 3 modules: 1) Finding region proposals, 2) detection model, 3) ideo-based TimeSformer

### A. Algorithm and Find Region Proposals

The algorithm used in module 1 is a combination of the methods presented in [31] and [32]. This method consists of three main parts. The first part selects the candidate areas in each image based on the chromatic features of the pixels. In the second part, the presence or absence of smoke or fire in the candidate areas of the first part is decided based on the dynamic features of each pixel at the time.

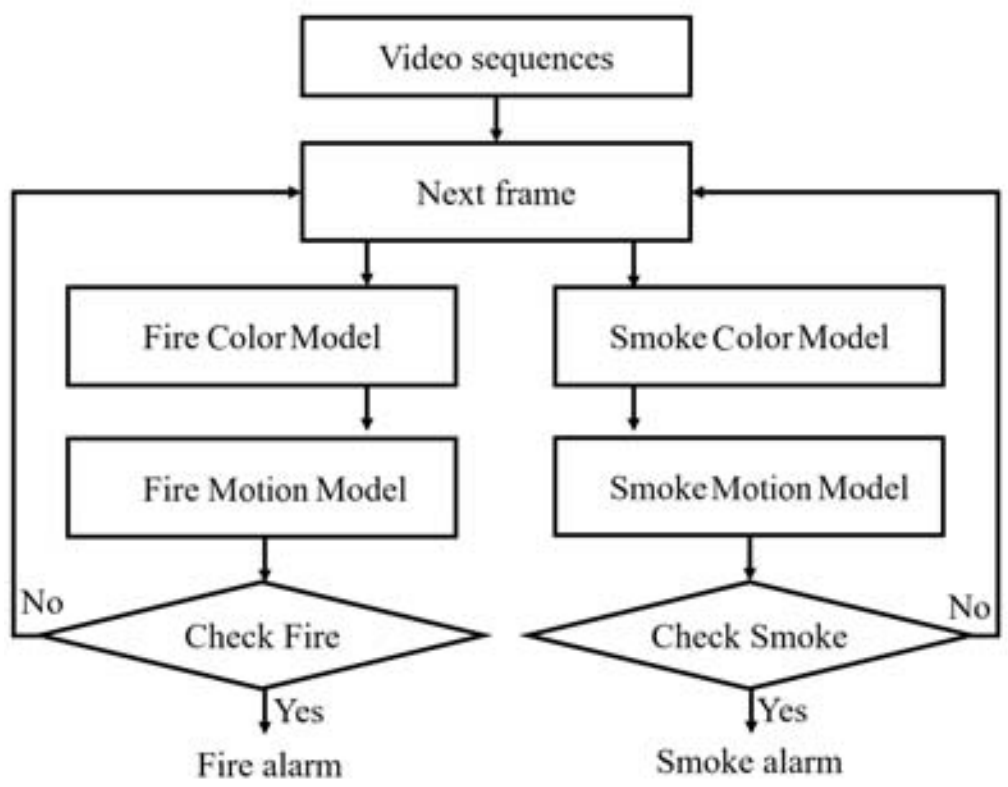

Fig. 2 Flowchart of finding region proposals

After that, by combining the candidate areas, if the number of adjacent points is more than a threshold value, a smoke or fire alarm is raised and the second module is executed. The flowchart of module 1 is as Fig. 2.

### B. Detection Network

In order to detect smoke and fire in the image, we use the yolov5 [33] detection network. This network has a good performance in the field of image-based object detection and has been used in many applications.

The architecture of the yolov5 network is given in Fig. 3. As shown in this figure, this network consists of three main parts. These three parts include the backbone, the neck, and the head.

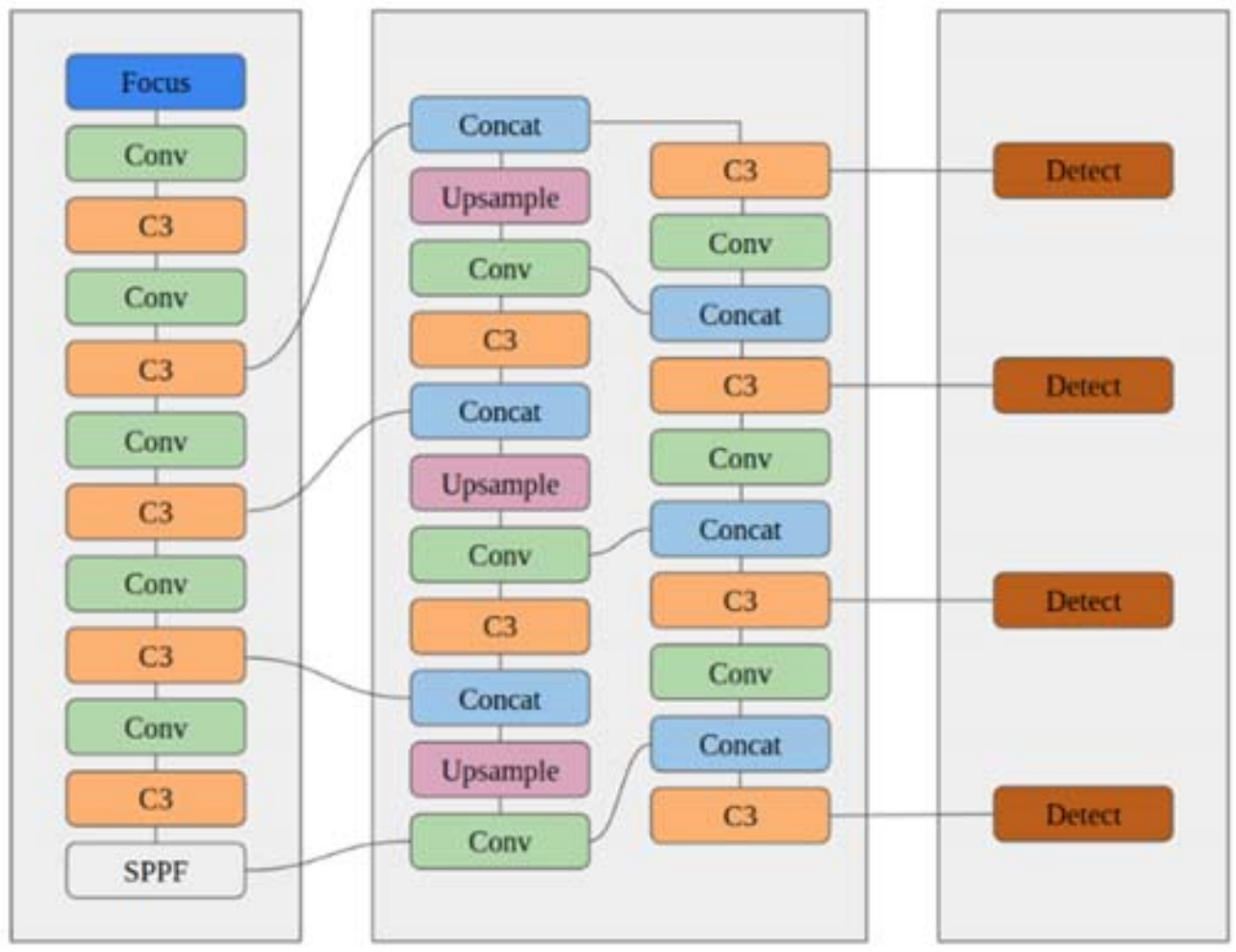

Fig. 3 Architecture of Yolov5m6 detection model

The model's backbone is used to extract important features of the input image. The yolov5 uses CSP (cross-stage partial) networks as the backbone. In this module, by eliminating the problem of repeated gradient, in addition to improving accuracy, the inference time is also increased.

The neck of the yolov5 network uses the PANet (path aggregation network), which improves the speed of data flow from the backbone to the network's head, as well as combining features with different scales. The second feature enables the model to detect objects with different dimensions, which improves the performance of the model. In the model used in this paper, four sets of features with different dimensions are used.

Finally, the network's head, called the YOLO layers, produces four outputs with different dimensions of the feature map that distinguish objects with different dimensions from small to large. Finally, from these feature maps, the location and category of objects in the image are obtained.

### C. Spatiotemporal Network

Since the detection model works on images, using the yolov5 detection model alone can cause false alarms in cases such as the sun, street lights, and so on. Some of these images along with the output of the detection network are given in Fig. 4. For this reason, it is necessary to use a model that measures the behavior of detected areas in consecutive frames.

For this purpose, the TimeSformer network [34], which was recently provided by Facebook AI researchers, has been used to classify videos. The TimeSformer architecture is based on the self-attention mechanism used in transformer models. In order to apply this structure to video, in this model, the input video is considered as a time-space sequence of image patches. The model derives the conceptual features of each patch by comparing it with other video patches. The computational complexity of 3D convolution-based architectures is high. This is due to the need to slide a large number of filters over all space-time locations in the video, while the TimeSformer is structured in such a way that its computational complexity is reduced. This is done by converting the video to non-overlapping patches and applying a kind of self-attention that prevents comparisons between all the patches.

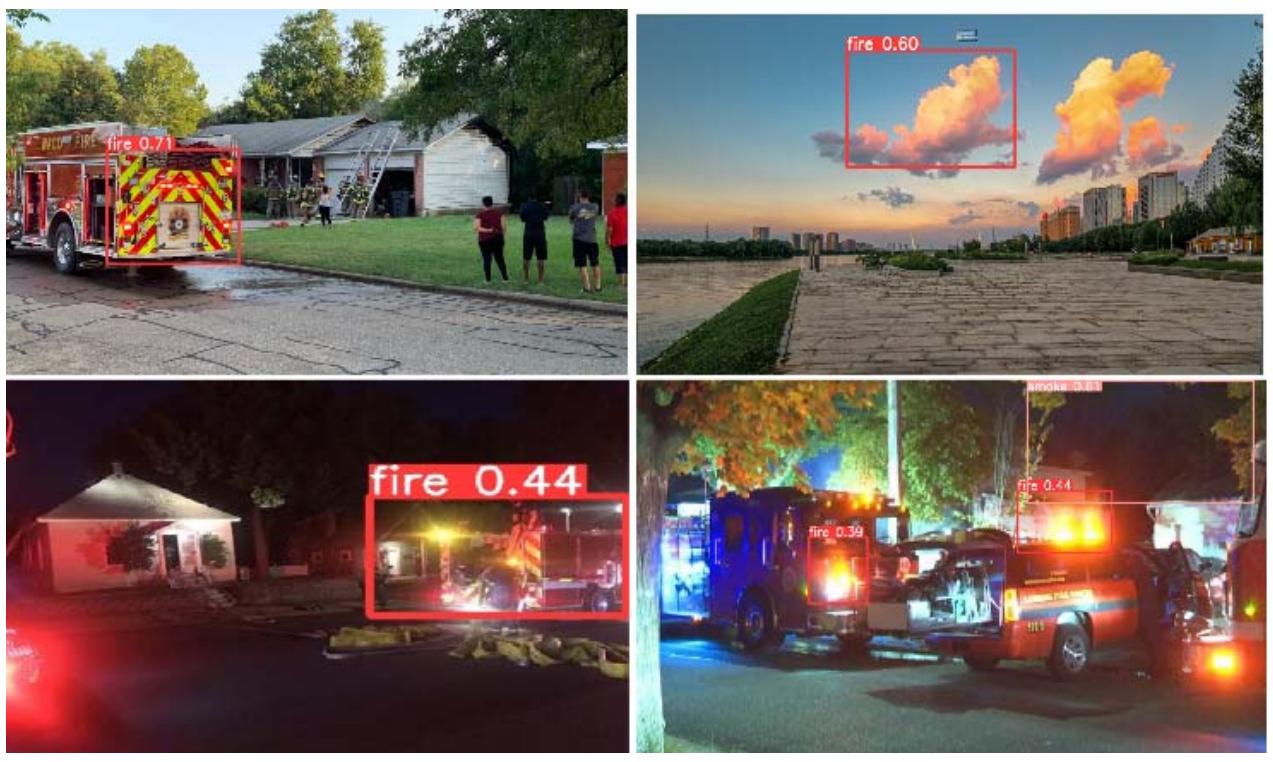

Fig. 4 Examples of detection model's false alarms

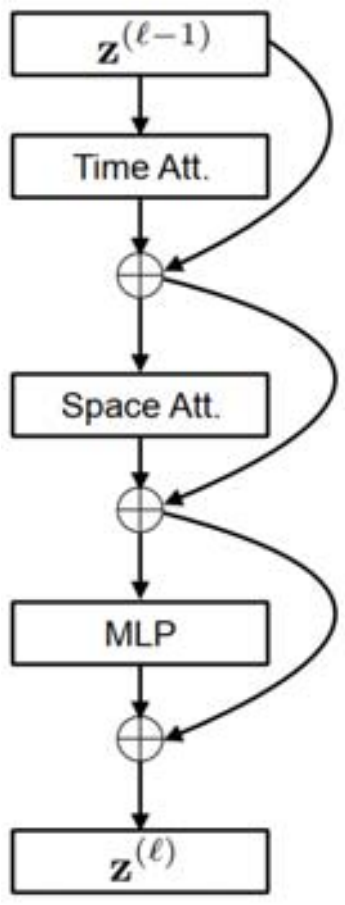

Fig. 5 The main submodule of TimeSformer

This structure is called divided space-time attention, and the idea is to apply temporal and spatial attention separately, one after the other. Also, the performance of this model has been improved in terms of accuracy compared to 3D convolution models.. The structure of submodules of the TimeSformer is given in Fig. 5. The input of the model $X$ is a video with dimensions $H \times W \times 3 \times F$, which includes $F$ $RGB$ frames with spatial dimensions $H \times W$ that are sampled from the original video. Each of these frames is divided into $N$ non-overlapping $P \times P$ patches. Each of these patches is then linearly mapped into the embedding vectors $z(0)$ with dimensions $D$ by a trainable matrix $E$ with dimensions $D \times 3P^2$:

$$z_{(p,t)}^{(0)} = Ex_{(p,t)} + e_{(p,t)}^{pos} \quad (1)$$

where $(p, t)$ indicates the space and time of each patch ($p = 1,2, \dots, N$ and $t = 1,2, \dots, F$). Index 0 means the first embedding and $e_{(p,t)}^{pos}$ is a location-dependent learnable embedding to encode the space-time location of each patch. The embedding $z$ is the input of TimeSformer.

The transformer consists of $L$ encoding blocks, in each block $l$, a vector of the value, key, and query of the sequence for each patch of $z(l-1)$ is calculated according to (2):

$$\begin{aligned} q_{(p,t)}^{(l,a)} &= W_Q^{(l,a)} LN\left(z_{(p,t)}^{(l-1)}\right) \in \mathbb{R}^{D_h} \\ k_{(p,t)}^{(l,a)} &= W_K^{(l,a)} LN\left(z_{(p,t)}^{(l-1)}\right) \in \mathbb{R}^{D_h} \\ v_{(p,t)}^{(l,a)} &= W_V^{(l,a)} LN\left(z_{(p,t)}^{(l-1)}\right) \in \mathbb{R}^{D_h} \end{aligned} \quad (2)$$

where $LN$ is the norm layer. The self-attention weights for the patch $(p, t)$ are then calculated as (3):

$$a_{(p,t)}^{(l,a)} = SM\left(\frac{q_{(p,t)}^{(l,a)T}}{\sqrt{D_h}} \left[k_{(0,0)}^{(l,a)} \left\{k_{(p',t')}^{(l,a)}\right\}_{\substack{p'=1,\dots,N \\ t'=1,\dots,F}}\right]\right) \quad (3)$$

where SM denotes the Softmax activation function. Direct use of these relationships results in quadratic computational complexity with respect to video dimensions and the number of frames. For this reason, the proposed method uses a separate calculation for space and time to reduce the calculations. Then the weighted sum of the value vectors with the coefficient obtained in (2) and (3) is obtained as (4):

$$s_{(p,t)}^{(l,a)} = a_{(p,t),(0,0)}^{(l,a)} v_{(0,0)}^{(l,a)} + \sum_{p'=1}^{N} \sum_{t'=1}^{F} a_{(p,t),(p',t')}^{(l,a)} v_{(p',t')}^{(l,a)} \quad (4)$$

Then by concatenating all these vectors and passing it through an MLP layer, the output is calculated as (5) and (6):

$$z'^{(l)}_{(p,t)} = W_0 \begin{bmatrix} s_{(p,t)}^{(l,1)} \\ \vdots \\ s_{(p,t)}^{(l,A)} \end{bmatrix} + z_{(p,t)}^{(l,a)} \quad (5)$$

$$z_{(p,t)}^{(l)} = MLP\left(LN\left(z'^{(l)}_{(p,t)}\right)\right) + z'^{(l)}_{(p,t)} \quad (6)$$

Finally, the classification of the final block is obtained using an MLP layer.

*D.Model Evaluation and Training Process*

Tuning and training of each module are done separately. The first module requires tuning the various threshold values, which is done by running the algorithm on different videos. Detection model is yolov5m6 with an input resolution of 1280×1280 with a learning rate of 0.001 and batch size 8 and TimeSformer model with an input resolution of 96×96, number of frames 16 with frame hop 8 and a measurement rate of 0.001 and Adam optimizer function.

To evaluate the accuracy performance of the model, the parameters of precision, recall and accuracy have been used, which are in accordance with (7)-(9). Our inference time results are based on NVIDIA GeForce GTX 1080 Ti hardware.

$$Preciosion = \frac{TP}{TP+FP} \quad (7)$$

$$Recall = \frac{TP}{TP+FN} \quad (8)$$

$$Accuracy = \frac{TP+TN}{Total} \quad (9)$$

where TP indicates true positive, FP indicates false positive, FN indicates false negative, and total indicates total results.

## IV. Numerical Results

### A. Datasets

This section describes the dataset preparation process. In this regard, a large number of images and videos were collected. In order to customize the model training dataset, these images and videos were manually reviewed, and among more than 50,000 images and 1000 videos, 3,500 images with about 4000 instances for fire and smoke, and 4500 videos were selected and labeled.

This dataset is used to train the smoke and fire detection and classification models, which are the second and third modules of the proposed method. Fig. 6 and Table I show the number and the distribution diagram of fire and smoke in the images. In Fig. 7 some examples of the collected dataset for module 2 are shown.

The rule to remove unusable images among the collected images was that the images with large or irregular fires, which were a problem for us in labeling, were removed. In addition, images without fire and smoke were added to the model training dataset, so that the model can use these images to distinguish between fire and smoke with objects such as sun and clouds. These images have been used to train the yolov5 model.

TABLE I
Number of Smoke, Fire, and Background Instances in Train and Test Detection Dataset

| Class | Train | Test |
|---|---|---|
| Fire | 3861 | 929 |
| Smoke | 4153 | 1068 |
| Background | 1876 | 407 |

To train the video classification model, due to the fact that there was no dataset including a large number of fire or smoke videos that were labeled, a large number of videos were collected and among them, the areas related to fire and smoke were labeled.

The videos were labeled in such a way that the fire or smoke areas in the video were cut and saved as a video for five seconds. This dataset included more than 17,000 videos, after a complete review and removing unusable videos and classifying them into three classes (Fire, Smoke, and Background), 1500 videos per class were obtained, which were used in training the final model.

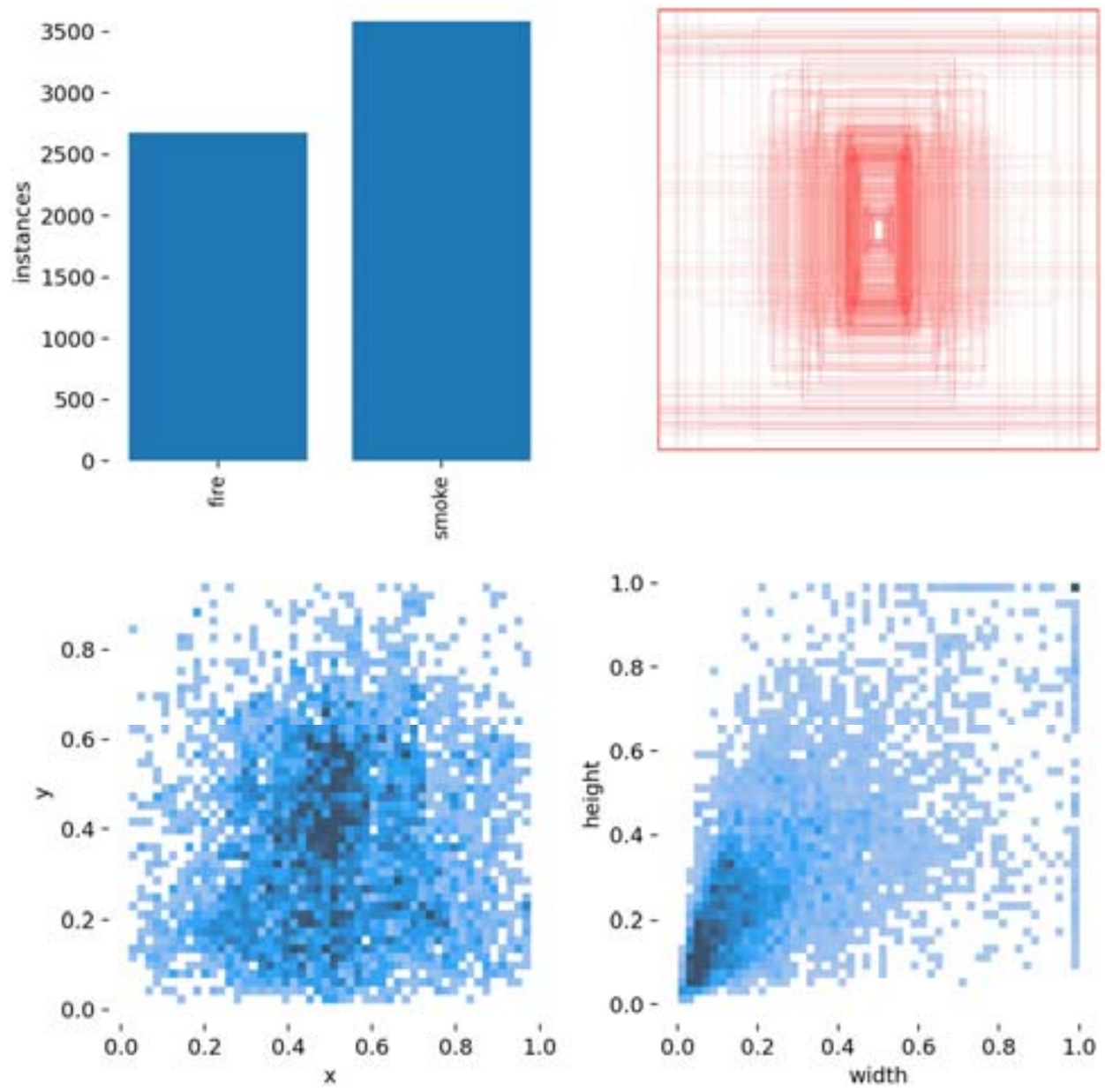

Fig. 6 Distribution diagrams of number and dimensions of fire and smoke detection dataset

### B. Performance

In this section, we compare the results of our proposed method with and without the third module and also compare it with one of the leading previous works FireNet [2].

Table II and Fig. 9 show the results obtained from the first two modules. For example, Table II shows that the accuracy of smoke detection with an IoU threshold value of 0,5 is 0.75 and the model inference time speed is 20 milliseconds per frame. Also, for example, Fig. 8 shows that the smoke detection error is of type false-negative and in fact 0.25 of smoke objects in the images are not detected.

TABLE II
Accuracy and Inference Time of Detection Model

| | YOLOV5m6 |
|---|---|
| Input Resolution | 1280×1280 |
| Fire Detection Accuracy | 0.92 |
| Smoke Detection Accuracy | 0.75 |
| MAP@0.5 | 0.83 |
| Inference Time (msec) | 20 |

The result of our proposed method improved after applying the video classification model, which is shown in Figs. 9-11 and Table III. Fig. 9 shows the precision vs recall diagram for different threshold values for the model with the third module, the model without the third module and the FireNet[2].

The area below the diagram shows the overall accuracy of the model, which shows that the performance of the model with the third module is better than the other two methods. Fig. 10 shows the confusion matrix of the model with the third module, which, as can be seen, has improved the performance of the model with the third module compared to the model without it (Fig. 8).

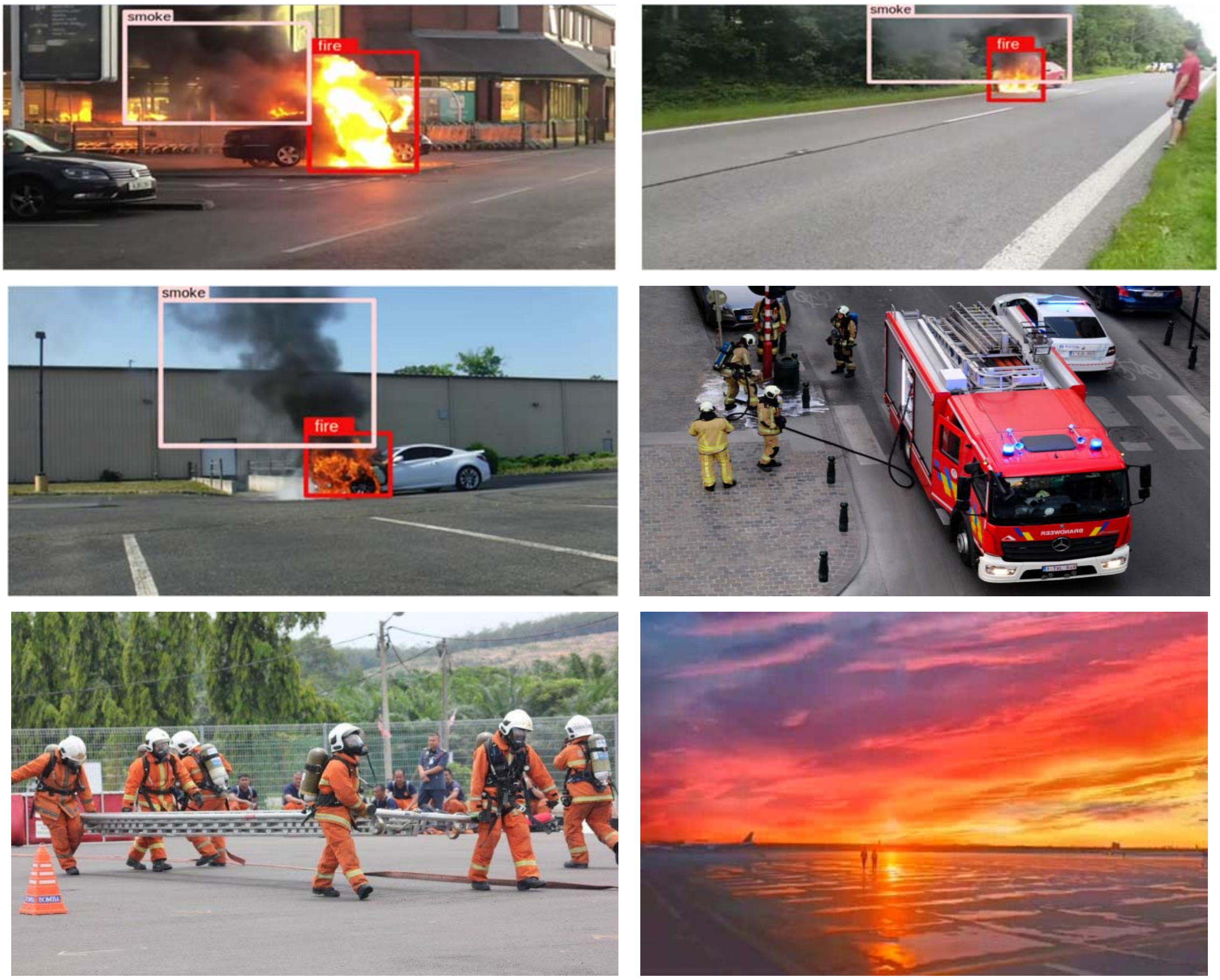

Fig. 7 Some examples of our collected detection dataset

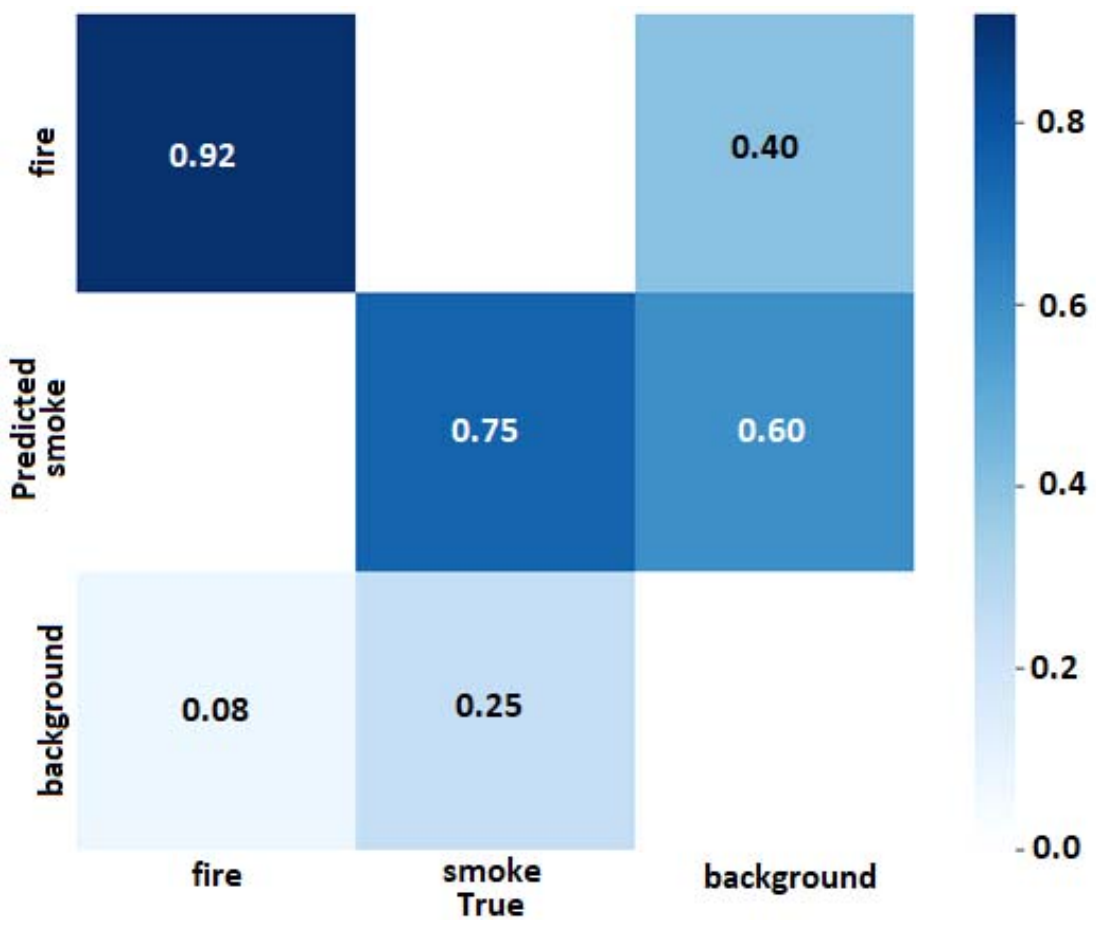

Fig. 8 Confusion matrix of our first two modules

An example of the performance of the third model is given in Fig. 9. In this figure, each box in the image represents the objects detected by the detection model. In the first line of each box, the output of the detection and in the second line, the output of the third module and the probability of the desired category are written. The color of each box is selected based on the output of the third module. As shown in the image, for example, the firefighter in the image was detected as fire by the detection model and then corrected by the third module. Also, the category of correctly identified boxes has not changed. This indicates the proper performance of the video-based TimeSformer model.

An example of the general test results of all models and algorithms on video is as Fig. 11.

To compare the accuracy of the model with similar models, the accuracy of the FireNet model on the database we use is as Table III.

## V. Conclusion

In this paper, a video-based spatiotemporal method for detecting smoke and fire is presented. This model consists of three main parts, each task was to reduce the error of the previous model so that the final model has a robust performance. The numerical results show that the proposed model performs better in terms of accuracy and inference time in comparison to other common methods. This is achieved by using a video-based model after an image-based detection model that simultaneously takes into account spatial and temporal characteristics.

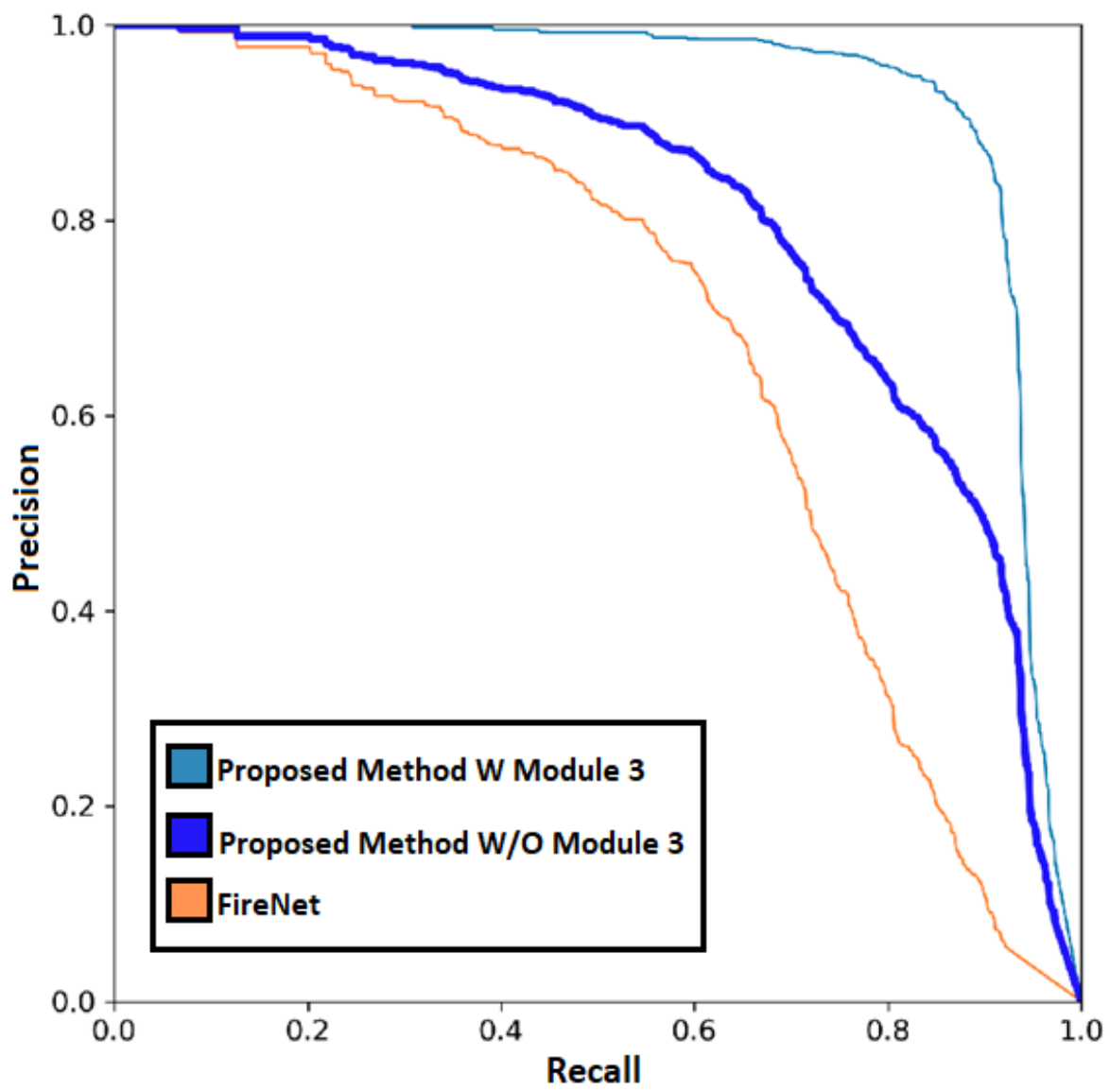

Fig. 9 Comparison of our method with and without module 3 and FireNet based on Precision vs Recall diagram

TABLE III
COMPARISON OF OUR PROPOSED METHOD WITH AND WITHOUT THE THIRD MODULE AND FIRENET

| Class | Fire | Smoke | Total |
|---|---|---|---|
| Proposed Method W/O Module 3 | 0.92 | 0.75 | 0.83 |
| Proposed Method W Module 3 | 0.95 | 0.85 | 0.80 |
| FireNet | 0.58 | 0.46 | 0.52 |

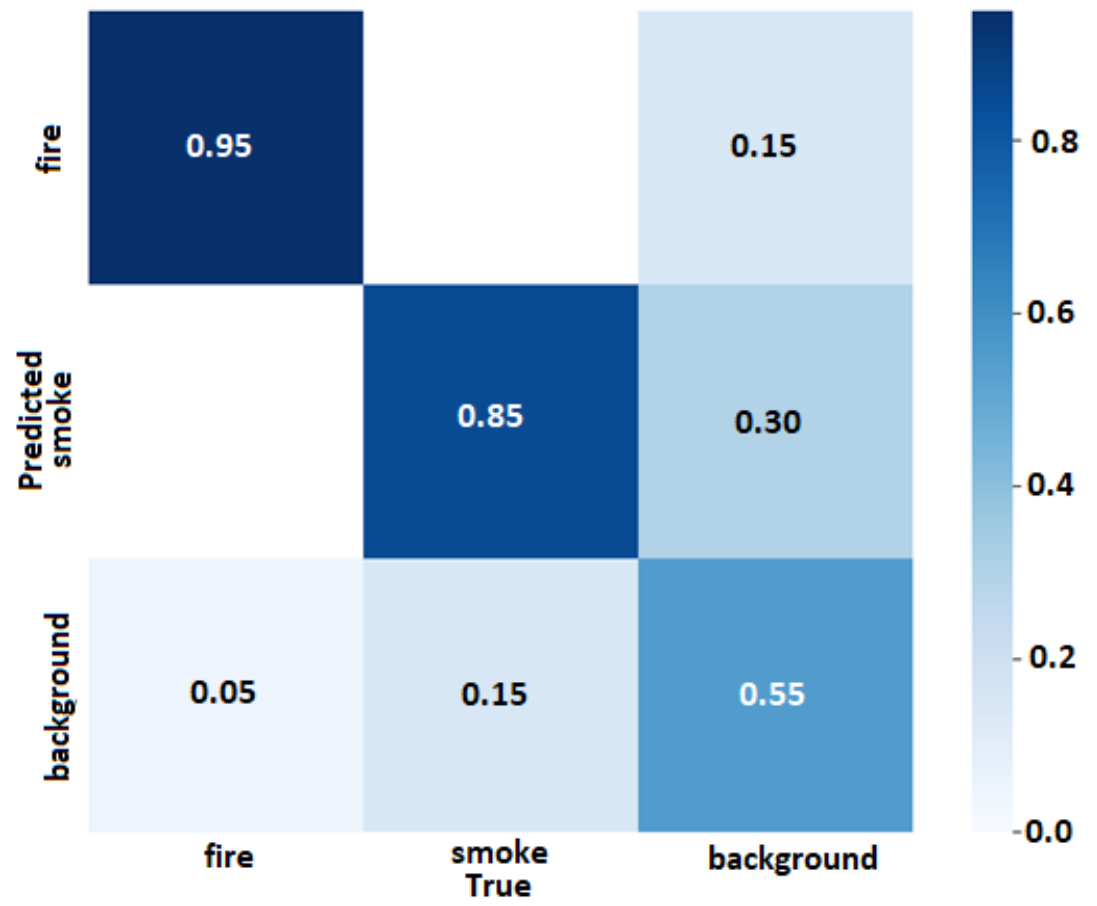

Fig. 10 Confusion matrix of our full approach

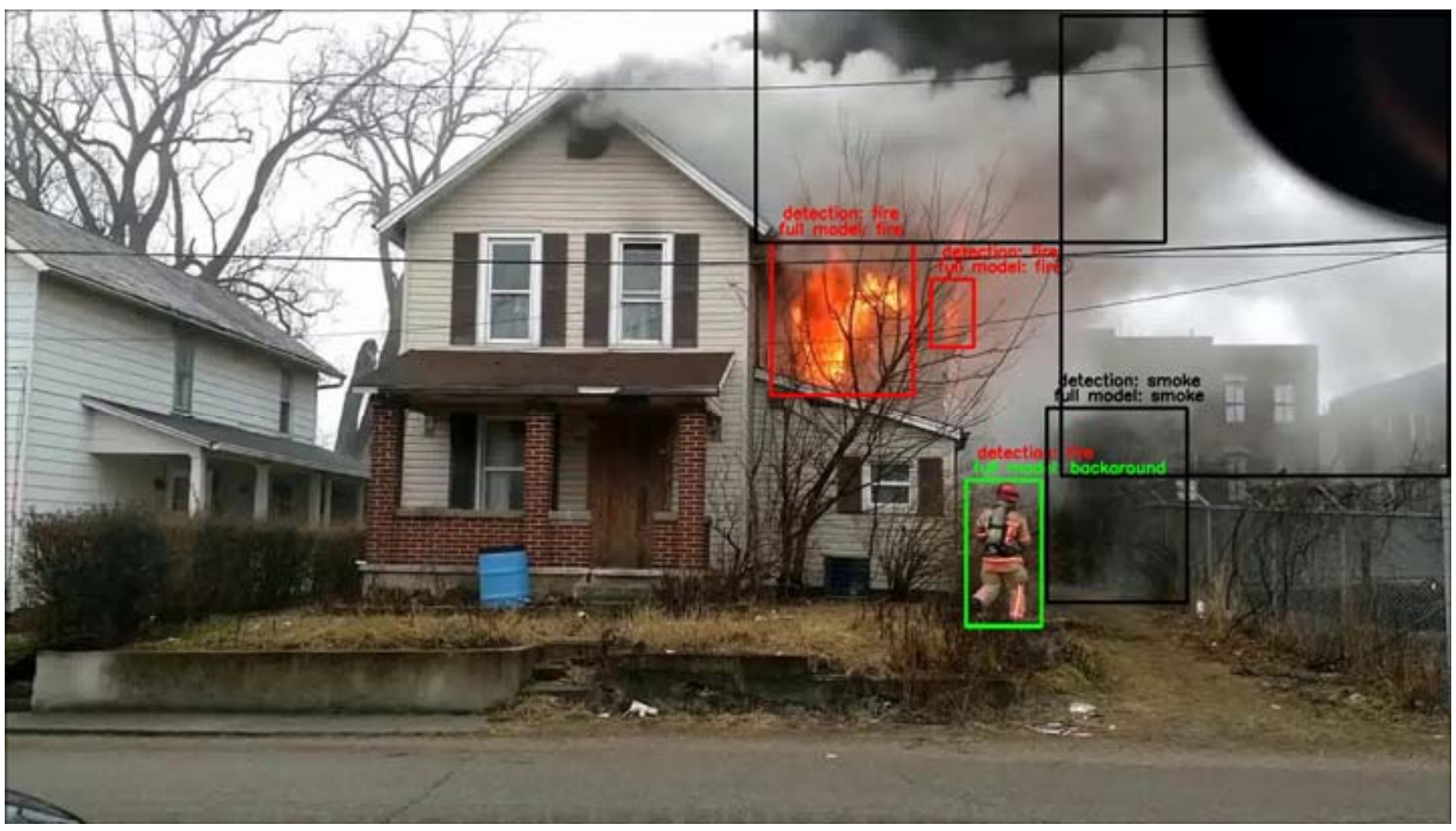

Fig. 11 An example of false alarm correction of the first two modules with the third module

ACKNOWLEDGMENT

This research was supported by Veunex. We thank our colleagues from AI department and health and safety team who provided insight and expertise that greatly assisted the research.